\documentclass[10pt]{article}
\usepackage[english]{babel}
\usepackage[utf8]{inputenc}

\usepackage{multicol}

\usepackage{tabu}
\usepackage[a4paper, total={6in, 7in}]{geometry}

\usepackage{amsmath}
\usepackage{amssymb}

\numberwithin{equation}{section}

\begin{document}

\markboth{\hfill{\rm B.\ Kerbl} \hfill}{\hfill {\rm The Impact and Outlook of 3D Gaussian Splatting \hfill}}

\title{The Impact and Outlook of 3D Gaussian Splatting}

\author{Bernhard Kerbl}

\maketitle

\begin{abstract}
Since its introduction, 3D Gaussian Splatting (3DGS) has rapidly transformed the landscape of 3D scene representations, inspiring an extensive body of associated research. Follow-up work includes analyses and contributions that enhance the efficiency, scalability, and real-world applicability of 3DGS. In this summary, we present an overview of several key directions that have emerged in the wake of 3DGS. We highlight advances enabling resource-efficient training and rendering, the evolution toward dynamic (or four-dimensional, 4DGS) representations, and deeper exploration of the mathematical foundations underlying its appearance modeling and rendering process. Furthermore, we examine efforts to bring 3DGS to mobile and virtual reality platforms, its extension to massive-scale environments, and recent progress toward near-instant radiance field reconstruction via feed-forward or distributed computation. Collectively, these developments illustrate how 3DGS has evolved from a breakthrough representation into a versatile and foundational tool for 3D vision and graphics.
\end{abstract}

\setcounter{tocdepth}{1}
\tableofcontents

\pagebreak

\section{Introduction}

In recent years, learnable scene representations have fundamentally reshaped how three-dimensional environments are captured, reconstructed, and rendered. Among these methods, 3D Gaussian Splatting (3DGS) has emerged as one of the most influential breakthroughs, offering a compelling balance between visual fidelity, reconstruction efficiency, and real-time rendering performance \cite{kerbl3Dgaussians}. Introduced as an alternative to implicit volumetric representations such as NeRF \cite{10.1145/3503250}, 3DGS represents scenes as collections of anisotropic Gaussian primitives, each parameterized by position, covariance, view-dependent color, and opacity. This formulation---building on the theory of elliptical weighted average splatting (EWA, \cite{10.5555/601671.601674})---enables direct rasterization through GPU-friendly procedures, achieving high-quality results at interactive frame rates while avoiding the expensive ray marching required by volumetric techniques. Sec.~\ref{sec:summary} revisits the mechanics of 3DGS.

The impact of 3DGS has been immediate and far-reaching. Within months of its release, numerous extensions and reinterpretations emerged across both academia and industry, seeking to adapt, generalize, and optimize the framework for new domains. However, the scope of this research landscape is now so vast that a single comprehensive review of its impact has become infeasible. The latter chapters of this report focus on selected advances that vividly exemplarize the trajectory of 3DGS and its ongoing integration into the broader ecosystem of 3D computer vision and graphics.

We begin by examining approaches that make 3DGS feasible under constrained computational resources, addressing scenarios such as training on commodity hardware, deployment on edge devices, and compression of Gaussian primitives (Sec.~\ref{sec:resources}). These works are central to democratizing the method, making real-time radiance field rendering accessible to a wider range of applications.

A second line of research explores dynamic 3D Gaussian Splatting, often referred to as 4DGS, where time-varying scenes or articulated objects are modeled through temporally coherent Gaussian representations (Sec.~\ref{sec:dynamic}). Such methods enable the capture and playback of dynamic performances, making them directly relevant to digital humans, telepresence, and immersive media.

In parallel, researchers have investigated the mathematical intricacies of 3DGS, particularly its appearance model, splatting formulation, and differentiable rendering properties (Sec.~\ref{sec:math}). These studies have provided deeper theoretical understanding and variants that improve accuracy, stability, and gradient propagation.

Another important branch of work targets practical deployment, bringing 3DGS to mobile and virtual reality (VR) platforms, where rendering efficiency and memory constraints are paramount (Sec.~\ref{sec:portable}). Combined with improvements in streaming and hardware integration, these efforts highlight the viability of Gaussian-based representations for real-world interactive systems.

Finally, recent research has demonstrated near-instant 3DGS reconstruction, leveraging feed-forward neural networks or distributed optimization pipelines to dramatically shorten reconstruction times (Sec.~\ref{sec:forward}). Such progress points toward the next stage of Gaussian-based methods—moving from offline capture to real-time scene acquisition.

Through this survey, we aim to contextualize 3DGS as a rapidly maturing paradigm that bridges the gap between learned 3D representations and real-time graphics. By highlighting these selected directions, we illustrate not only the versatility of 3D Gaussian Splatting but also its role in shaping the future of efficient, scalable, and accessible 3D scene capture and rendering.

\newcommand{\gs}[0]{3DGS~}

\section{Background} 
\label{sec:summary}

\gs heavily exploits the concepts proposed by the volume rendering equation:
\begin{equation}
C(\vec{r}) =  \int_0^t \vec{c}\left(\vec{r},t\right) \, \sigma\left(\vec{r},t\right) \, T(\vec{r},t) \,  dt, \hspace{1cm} T(\vec{r},t)=e^{-\int_0^t \sigma\left(\vec{r},s\right) \, ds },
\label{eq:volume}
\end{equation}
where $C(\vec{r})$ is the eventual color for a pixel along a given ray $\vec{r}$, $\sigma\left(\vec{r},t\right)$ models light attenuation and extinction strength throughout the field, and $c\left(\vec{r},t\right)$ is the view-dependent emitted radiance at a given point on the ray. 
To produce variations in the radiance fields, \gs represents a scene as a mixture of 3D Gaussians, each given by:
\[
G(\vec x) = e^{-\frac{1}{2}(\vec{x}-\vec{\mu})^T\Sigma^{-1}(\vec{x}-\vec{\mu})},\hspace{1cm}\Sigma = R S S^T R^T,
\]
where $\vec{\mu}$ is the Gaussian's mean, $R$ is a rotation matrix and $S$ is a diagonal scaling matrix. 
Combining these attributes enables to model a non-normalized Gaussian distribution with a given mean $\vec{\mu}$ and covariance $\Sigma$ in 3D space.
In practice, \gs considers all Gaussians to be non-overlapping, i.e., separated in space, effectively compressing their extent to a Dirac delta along the ray.
The Dirac delta of the Gaussian $i$ is located at 
\begin{equation}
    t_i = \vec \mu_i^T \vec v,
    \label{eq:viewspacedepth}
\end{equation} i.e., the projection of the mean $\vec \mu_i$ onto the view direction $\vec v$, independent of the individual ray $\vec r$.
Following the EWA framework, \gs then approximates the actual projection of each Gaussian via an orthogonal projection to construct a 2D \emph{splat} on screen $G_2$~\cite{10.5555/601671.601674}.
This discretization into non-overlapping primitives enables fast volume rendering via rasterization: Equation
\ref{eq:volume} becomes
\begin{equation}
C(\vec{r}) = \sum_{i=1}^{N} \vec{c}_i  \alpha_i \prod_{j=1}^{i-1}(1-\alpha_j),
\label{eq:volume_desc}
\end{equation}
where $i$ iterates over the $N$ Gaussians that influence the ray in the ordering of $t_i$, and $\alpha_i$ is the blending weight for the Gaussian during accumulation along the ray, i.e., $G_2(x,y)$, multiplied by a learned per-Gaussian \emph{opacity} value. During training, the attributes of representative Gaussians are optimized via gradient descent. Usually, additional Gaussians are incrementally introduced in areas that require more thorough sampling ("densification").

\section{3DGS with Limited Resources} 
\label{sec:resources}
3DGS can demand substantial memory and compute resources, which limits deployment on commodity, mobile, and web platforms. Recent work addresses this through compression of per-scene assets, training policies that bound model growth, and reducing runtimes with algorithmic optimizations.

\paragraph{Compression-focused methods.}
\emph{Reducing the Memory Footprint of 3DGS} combines (i) pruning of redundant splats, (ii) adaptive reduction of view-dependent appearance parameters (e.g., SH order), and (iii) quantization of per-Gaussian attributes to shrink storage while preserving visual quality \cite{10.1145/3651282}.
\emph{Compressed 3D Gaussian Splatting for Accelerated Novel View Synthesis} encodes per-Gaussian properties using learned, compact representations (e.g., codebooks/latents) that are decoded at render time, reducing per-scene size and bandwidth \cite{Niedermayr_2024_CVPR}.
\emph{Compact 3D Gaussian Representation for Radiance Field} pursues a compact parameterization by coupling geometry/appearance compression (e.g., shared fields or tensors with vector quantization) with a rendering scheme compatible with 3DGS, targeting small on-disk assets with real-time playback \cite{10655367}.

\paragraph{Budget-aware training and spatial efficiency.}
\emph{EAGLES} reduces per-Gaussian storage by replacing heavy, per-point parameters with compact embeddings and trains under a coarse-to-fine schedule, together with redundancy control \cite{10.1007/978-3-031-73036-8_4}.
{\emph{Taming3DGS} enforces deterministic, budget-aware densification, keeping the primitive count bounded during optimization to meet fixed VRAM or runtime targets \cite{10.1145/3680528.3687694}.
Mini-Splatting improves spatial distribution of Gaussians via paired densification/simplification steps so that fewer, better-placed splats sustain reconstruction quality under strict budgets \cite{10.1007/978-3-031-72980-5_10}.
As an example of industry efforts towards low-resource 3DGS tooling, \emph{Brush} (Google Deepmind) demonstrates training and rendering of 3DGS in the browser using WebGPU, highlighting a path to zero-install demos and broader hardware coverage; this in turn encourages smaller models and lightweight encodings.

\paragraph{Summary.}
Across these directions—compression of attributes, budgeted densification, spatial reorganization, and portable runtimes—the community is converging on compact, resource-efficient 3DGS pipelines that retain the real-time and high-fidelity strengths of the original formulation.

\section{Dynamic 3D Gaussian Splatting}
\label{sec:dynamic}
Classical 3D Gaussian Splatting assumes a static world: each Gaussian is fixed in space, and rendering is purely a function of viewpoint. This assumption breaks immediately in realistic settings with nonrigid motion, human performance, scene flow, and long sequences. Extending 3DGS to dynamic scenes turns out not to be “just add time as a fourth coordinate,” because doing so raises three hard constraints simultaneously:

\begin{enumerate}
\item{
Temporal coherence} – primitives should persist and remain identifiable over time, rather than being re-fit from scratch every frame.
\item{
Real-time rendering of motion} – dynamic reconstruction should be renderable interactively, not just offline.
\item{Scalability in sequence length} – the representation must not explode in size for multi-second or minute-long videos.
\end{enumerate}

\paragraph{Dynamic 3D Gaussians: Tracking by Persistent Dynamic View Synthesis \cite{luiten2023dynamic}} reframes a dynamic scene not as a sequence of unrelated per-frame reconstructions, but as a single set of Gaussians with consistent appearance through time. Each Gaussian is allowed to move and rotate from frame to frame, but its appearance parameters (color, opacity, size) are constrained to remain constant, and its motion is regularized to be locally rigid. 
Conceptually, this paper establishes the idea that Gaussians are not just render primitives but scene elements with identity. That is the first pillar of dynamic 3DGS: persistence + motion, rather than per-frame respawning.

\paragraph{4D Gaussian Splatting for Real-Time Dynamic Scene Rendering \cite{Wu_2024_CVPR}} takes a different stance. Rather than starting from tracking, it asks: what is the minimal extension to the 3DGS pipeline that makes dynamic view synthesis itself real-time? The answer they propose is to lift Gaussians into a 4D representation, where time is treated as an explicit dimension alongside x,y,z, and to design a renderer that can evaluate those spatio-temporal Gaussians fast enough for interactive playback. 
Technically, the contributions are:
\begin{itemize}
\item
Deformable spatio-temporal Gaussians. Instead of maintaining an independent Gaussian cloud per frame, the method models a single dynamic field whose Gaussians can deform over time, capturing nonrigid motion. This avoids naïvely duplicating millions of primitives across frames. 
\item
Real-time splatting in 4D. The paper retains the splatting-style rasterizer of static 3DGS (i.e., GPU-friendly blending of anisotropic Gaussians) and adapts it so that interpolation over time and view can still be executed at interactive frame rates, demonstrated at tens of FPS on commodity GPUs. 
\item
Fast optimization. By treating the scene holistically rather than fitting per-frame models, they report training times on the order of tens of minutes for high-resolution dynamic captures, rather than many hours. 
\end{itemize}

Where Dynamic 3D Gaussians emphasized identity and physical consistency of individual primitives, 4D Gaussian Splatting emphasizes throughput: make dynamic neural rendering as fast and interactive as static 3DGS, without collapsing under motion complexity. The result is a practical real-time 4D renderer rather than per-frame reconstructions stitched together.

\paragraph{Representing Long Volumetric Video with Temporal Gaussian Hierarchy \cite{10.1145/3687919}} Both of the above methods assume clips of manageable duration. This third line of work asks what happens when “dynamic” means minutes of capture, not a few hundred frames.

The core observation is that dynamic scenes are not uniformly dynamic: some regions (background, static objects) barely change; others (hands, cloth, facial detail) evolve rapidly. The paper encodes this by building a multi-level temporal hierarchy of Gaussians, where higher levels capture stable structure that persists for long spans of time and can be reused across many frames.
Further, lower levels capture fast-changing, high-frequency local motion, but only where and when it is needed.
Two technical payoffs fall out of that hierarchy:
\begin{itemize}
\item
Temporal reuse instead of temporal duplication. Rather than instantiating a fresh copy of the entire scene’s Gaussians every frame, large parts of the scene are pointed to by reference across long intervals. That slashes the effective primitive count for long sequences. 
\item
Bounded working set. Because the hierarchy is tree-structured over time, the renderer (and the optimizer) can assemble only the subset of Gaussians active at a particular timestamp. This keeps peak GPU memory usage roughly constant even as the represented video length grows to thousands of frames.
\end{itemize}

\paragraph{Summary}
Taken together, these works outline what “dynamic 3DGS / 4DGS” currently means technically: a representation whose primitives (i) persist and move coherently in time, (ii) can be splatted at interactive rates for arbitrary viewpoints, and (iii) can be organized hierarchically so that even minute-long captures remain tractable. This progression turns 3DGS from a static scene capture method into a candidate backbone for editable, replayable, long-duration volumetric video.

\section{Mathematical Intricacies of 3DGS}
\label{sec:math}
In its seminal form, the 3DGS formulation altered several key aspects of the underlying volume rendering theory, directly impacting the appearance of Gaussians on screen: this includes noticeable aliasing artifacts for rescaled or intensely zoomed views, overall appearance modeling of Gaussians, as well as significant distortions of Gaussians in periphery zones where the orthogonal projection strongly deviates strongly from the correct projection behavior.

\paragraph{Antialiasing}
Several works address aliasing and sampling-scale mismatches in 3DGS. These appear most prominently when zooming out on a 3DGS scene, or the rendered image resolution differs strongly from training. \emph{Mip-Splatting} introduces 3D and 2D scale-aware filtering to produce consistent, alias-free results under zooming or resolution changes \cite{Yu2023MipSplatting}. \emph{Multi‑Scale 3D Gaussian Splatting for Anti‑Aliased Rendering} \cite{Yan2023MultiScale3G} introduces a multiscale representation of Gaussians---maintaining smaller splats for high-resolution views and larger ones for distant or low-resolution views—to reduce aliasing when camera scale changes. 
\emph{Anti-Aliased and Artifact-Free 3D Gaussian Rendering} \cite{steiner2025aaags} builds on this further by applying an adaptive 3D smoothing filter to each Gaussian based on viewpoint sampling frequency, and by implementing view-space bounding and 3D tile culling that reduce popping and projection artifacts in out-of-distribution views. 

Together, these studies clarify that aliasing in 3DGS is not just a texture-filter issue—it arises from mismatches between Gaussian footprints, viewing scale, and sampling resolution.

\paragraph{3DGS Appearance Model}
Two recent papers \cite{10.1111:cgf.70032, talegaonkar2025vol3dgs} independently tackle the assumption that rasterized splats (with alpha blending) sufficiently capture light transport. These works clarify when and why the simplifying appearance assumptions of the 3DGS appearance model hold or fail. Celarek et al.\ comparatively evaluate full volumetric integration versus the typical rasterization used in 3DGS, and show that, while full volumetric treatment can improve quality in sparse-splats regimes, the efficient rasterization regime of 3DGS remains competitive in high-splats settings, with markedly lower compute effort. 

\paragraph{Distortion Error}
Projection and geometric distortion of Gaussians---especially in wide-angle or peripheral views---is another subtle source of artifacts in 3DGS. A first-order error analysis of Gaussian projection under screen-space assumptions derives an improved projection scheme that aligns Gaussian covariance tensors to the view tangent plane and reduces systematic distortion \cite{10.1007/978-3-031-72643-9_15}. In parallel, \emph{Don’t Splat Your Gaussians} and \emph{3DGUT} \cite{10.1145/3711853, wu20253dgut}, as well as Hahlbohm et al.~\cite{hahlbohm2025htgs}, propose rendering adjustments and implicit correction layers that mitigate the residual distortion effect in rendering. These works highlight that alignment between Gaussian primitve geometry and projection geometry is critical to high-quality novel-view synthesis.

\paragraph{Summary}
Collectively, these mathematical investigations deepen the theoretical foundation of 3DGS. They show that for robust performance and artifact-free rendering, one cannot treat Gaussian splats as naïve 2D patches: anti-aliasing requires scale-aware filtering and bounding, appearance modeling must consider volumetric vs raster trade-offs, and projection geometry must align with primitive covariance. These insights anchor the next generation of 3DGS pipelines that aim for both real-time performance and high-fidelity results.

\section{3DGS for Virtual Reality} 
\label{sec:portable}
Although 3DGS provides high-fidelity and real-time rendering rates on dedicated machines (e.g., desktop computers), weaker hardware can easily struggle with splatting millions of Gaussian primitives. Wearable VR devices, in particular, face elevated challenges due to their demand for stereoscopic presentation (i.e., inherent doubling of rendering load) and the fine-grained retinal coverage (i.e., display resolution) required to provide crisp visuals. The situation is further complicated by the fact that the usually subtle artifacts of 3DGS are exacerbated by the wide field of view and the high-frequency camera motions associated with VR. Hence, targeted solutions are required that can ensure both artifact-free rendering behavior \emph{and} high performance.

\paragraph{VR‑Splatting: Foveated Radiance Field Rendering via 3D Gaussian Splatting and Neural Points \cite{10.1145/3728302}} addresses the core VR challenge of rendering high‐fidelity scenes at interactive framerates by leveraging the human visual system’s gaze‐dependency. Technically, the authors combine a high‐resolution neural point rendering subsystem for the gaze (foveal) region with a lower‐overhead 3DGS subsystem for the periphery. By foveating the workload, they reduce per‐frame computation without compromising perceived sharpness. They further analyse the trade-offs of existing novel-view methods (NeRF, pure 3DGS) under VR latency constraints and adapt the pipeline accordingly. In effect, they demonstrate that the hybrid renderer achieves both real-time performance and high perceived fidelity in VR—pointing toward a practical path for 3DGS in head‐mounted displays. 

\paragraph{VRSplat: Fast and Robust Gaussian Splatting for Virtual Reality \cite{tuposter, 10.1145/3728311}} takes a more systems‐centric approach: instead of just reallocating detail (as the previous paper does), it refines the 3DGS pipeline itself to meet the demands of VR (high framerate stereo, large field of view, head motion). Key technical contributions include eliminating temporal “popping” by stabilising depth and visibility transitions, correcting projection distortions (especially in peripheral vision) by integrating an improved projection model, and implementing a foveated rasteriser that runs a single GPU pass for focus and periphery—thus maximising GPU utilisation. The result is a VR‐class 3DGS engine that reports 72+ fps in stereo and eliminates objectionable artefacts for head-mounted display use. 

\paragraph{Summary}
These works demonstrate the growth of 3D Gaussian Splatting in VR contexts: one by embedding perceptual awareness (foveation + hybrid representation), the other by engineering rendering stability and performance at the system level (latency, stereo, projection). They mark a transition from “3DGS for static novel-view synthesis” to “3DGS for immersive VR experiences.”

\section{Toward Instant 3DGS Reconstruction} 
\label{sec:forward}

A key frontier in the evolution of 3D Gaussian Splatting (3DGS) is achieving scene-ready reconstructions in seconds or less, rather than the minutes to hours typical of earlier pipelines. Here we review four representative methods, each interpreting “instant” in a slightly different way: feed-forward generalization to new scenes, very sparse input reconstructions, unposed image sequences, and live-event streaming.

\paragraph{PixelSplat \cite{charatan23pixelsplat}} proposes a feed-forward architecture that, given just a pair of posed images, produces a full 3D Gaussian splatting scene ready for rendering. The technical novelty lies in (i) predicting a dense probability distribution over 3D space for Gaussian mean placement (avoiding iterative fitting), and (ii) sampling that distribution via a differentiable reparameterization trick, allowing end-to-end learning of Gaussian placement, shape, and appearance. 

At inference the model bypasses per-scene optimization entirely: the network predicts the Gaussian parameters directly, enabling reconstructions in “real-time” (order milliseconds) for small scenes.

\paragraph{GS-LRM \cite{gslrm2024}} takes the “instant” ambition further: from 2-4 posed images, the transformer‐based model predicts dense per-pixel Gaussian primitives in $\approx$0.23 seconds on a single A100 GPU. Its contribution includes processing multiple views via patchified tokens, decoding directly to Gaussian parameters (centroid, covariance, appearance), and doing so in a single feed-forward pass rather than an iterative solver. This enables high‐quality reconstructions of complex scenes with large scale variations practically “on demand.” Importantly, GS-LRM interprets “instant” as very sparse input + very fast inference rather than long capture sequences.
\pagebreak
\paragraph{On‑the‑fly Reconstruction for Large‑Scale Novel View Synthesis from Unposed Images \cite{10.1145/3730913}.} Where the previous two assume at least partial calibration or posed input, this method handles unposed image streams and large‐scale scenes, producing usable Gaussian splatting reconstructions during capture. It does so by introducing (i) a fast pose-initialization module using learned features and GPU-friendly mini bundle adjustment, and (ii) direct incremental spawning and clustering of Gaussian primitives to avoid full offline optimization. Here “instant” means near-real-time reconstruction from previously unprocessed inputs, enabling novel-view synthesis immediately after capture ends.

\paragraph{Echoes of the Coliseum: Towards 3D Live Streaming of Sports Events \cite{10.1145/3731214}.}
This study addresses live, \emph{dynamic}, and volumetric content—e.g., sports events—and frames “instant” as live streaming readiness: multi-camera input, distributed processing, coarse-to-fine actor/environment separation, and deferred Gaussian refinement permit interactive free-viewpoint exploration of a live scene. Technically it factors scene reconstruction into actor-centric tasks and static updates, harnesses geometric priors for speed, and pipelines Gaussian updates so that a near‐live scene representation becomes queryable during an event. The emphasis here is runtime and latency: transform capture into a viewable experience.

\paragraph{Summary}
These works represent the spectrum of “instant” reconstruction in 3DGS:
Through different techniques—feed‐forward prediction, transformer decoding, pose‐free input handling, distributed live pipelines—they push the 3DGS representation from the batch offline domain toward real-time. As a result, Gaussian splatting no longer remains “fast relative to NeRF”, but becomes practically instantaneous in a range of operating conditions.

\bibliographystyle{plain}  
\bibliography{bib}

\noindent Institute of Visual Computing \& Human-Centered Technology\\
TU Wien\\
1040 Vienna, Austria.\\
kerbl@cg.tuwien.ac.at}

\end{document}